\documentclass[11pt,anonymous]{article}
\usepackage{nodalida2025}
\usepackage{times}
\usepackage{url}
\usepackage{latexsym}
\usepackage{tabularx}
\usepackage{booktabs}
\usepackage{hyperref}
\usepackage{amsmath}
\usepackage{multirow}
\usepackage[utf8]{inputenc} % allow utf-8 input
\usepackage[T1]{fontenc}    % use 8-bit T1 fonts
\usepackage{hyperref}       % hyperlinks
\usepackage{amsfonts}       % blackboard math symbols
\usepackage{nicefrac}       % compact symbols for 1/2, etc.
\usepackage{microtype}      % microtypography
\usepackage{xcolor}         % colors
\usepackage{graphicx}
\usepackage{float} 
\usepackage{multicol}
\usepackage{amsmath}    % For math formatting
\usepackage{adjustbox} % To fit table without shrinking text

\usepackage{multirow}    % For multi-row cells
\usepackage{xcolor}      % For colored symbols
\usepackage{pifont}      % For tick and cross symbols
\usepackage{longtable} % For tables that span multiple pages
\usepackage{afterpage}

\newcommand{\cmark}{\textcolor{green}{\ding{51}}}%
\newcommand{\xmark}{\textcolor{red}{\ding{55}}}%

\renewcommand{\normalsize}{\fontsize{11}{13.2}\selectfont}

\aclfinalcopy % Uncomment this line for the final submission

\title{\fontsize{15}{18}\selectfont\bfseries LAG-MMLU:  Benchmarking Frontier LLM Understanding in Latvian and Giriama}

\author{
    Naome A. Etori$^{1,4}$, Kevin Lu$^3$, Randu Karisa$^4$ and Arturs Kanepajs$^2$ \\
    $^1$University of Minnesota-Twin Cities \\
    $^2$Independent Researcher \\
    $^3$Bellarmine College Preparatory \\
    $^4$Masakhane \\
    \texttt{etori001@umn.edu}
}

\date{}

\begin{document}
%\pagenumbering{gobble} 
\maketitle
\begin{abstract}
As large language models (LLMs) rapidly advance, evaluating their performance is critical. LLMs are trained on multilingual data, but their reasoning abilities are mainly evaluated using English datasets. Hence, robust evaluation frameworks are needed using high-quality non-English datasets, especially low-resource languages (LRLs). This study evaluates eight state-of-the-art (SOTA) LLMs on Latvian and Giriama using a Massive Multitask Language Understanding (MMLU) subset curated with native speakers for linguistic and cultural relevance. Giriama is benchmarked for the first time. Our evaluation shows that OpenAI's o1 model outperforms others across all languages, scoring 92.8\% in English, 88.8\% in Latvian, and 70.8\% in Giriama on 0-shot tasks. Mistral-large (35.6\%) and Llama-70B IT (41\%) have weak performance, on both Latvian and Giriama. Our results underscore the need for localized benchmarks and human evaluations in advancing cultural AI contextualization.
\end{abstract}

\section{Introduction}
\begin{figure}[ht]
    \centering
    \includegraphics[width=0.85\columnwidth]{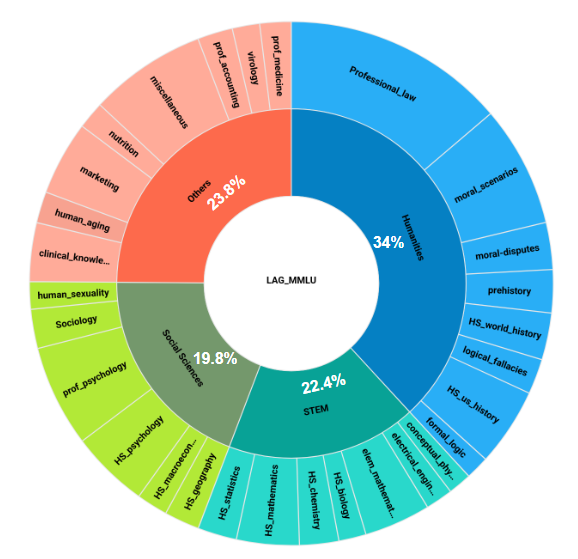} % Slightly smaller than full column width
    \caption{Distribution of subjects and categories of the subjects in LAG-MMLU. We included subjects with a frequency of 7 and above. 55 out of 57 subjects are represented. See more details in Table~\ref{tab:subject_distribution} in the appendix.}
    \label{fig:lag_mmlu_distribution}
\end{figure}

% \begin{figure}[ht]
%     \centering
%     \includegraphics[width=\columnwidth]{SUNBURST2.PNG} % Adjusts to fit one column
%     \caption{Distribution of subjects and categories of the subjects in LAG-MMLU. We included subjects with a frequency of 7 and above. 55 out of 57 subjects are represented.}
%     \label{fig:lag_mmlu_distribution}
% \end{figure}

Large Language Models (LLMs) have significantly advanced natural language processing (NLP), enabling progress in tasks such as machine translation(MT) \cite{wang2023afrimte}, knowledge reasoning \cite{singh2024global}, and Conversational \cite{etori2024wiscompanion}. Models like GPT-4 \cite{hurst2024gpt, OpenAIO1SystemCard2024}, LLaMA \cite{touvron2023llama, dubey_llama_2024}, and BLOOM \cite{le2023bloom} have demonstrated remarkable performance, attributed mainly to their pretraining on vast, multilingual datasets. Therefore, it is crucial to rigorously assess LLMs' ability to represent knowledge and reason across multiple languages.\cite{koto2023large, adelani2024irokobench}.

Benchmarking is critical in assessing LLM's performance and capabilities across diverse tasks due to its standardized evaluation frameworks. However, the limited availability of robust multilingual benchmarks has led to the continued focus on a few common and widely spoken languages in LLMs and within the NLP research community \cite{dargis2024evaluating, li2023cmmlu}. 

As a result, some existing frameworks, such as Massive Multitask Language Understanding (MMLU) \cite{hendrycks2020measuring}, are widely recognized as gold standards for evaluating the capabilities of large language models (LLMs). MMLU assesses knowledge across various subject domains, including STEM, humanities, and social sciences.  For instance, MMLU was recently featured in the Gpt-4o system card \cite{hurst2024gpt}. Due to its reliance on U.S. English, its applicability for assessing multilingual capabilities is limited unless the dataset is translated into different languages. However, translating English benchmarks introduces biases and noise and fails to account for region-specific and cultural understanding. This may also add another layer of cultural misalignment of the target language, such as questions about U.S. laws or customs \cite{liu2023multilingual}.

Recently, NLP researchers have addressed this limitation; several diverse linguistic and cultural context MMLU evaluation benchmarks have been developed to address the gap, such as ArabicMMLU \cite{koto2024arabicmmlu} evaluates tasks from North Africa, the Levant, and the Gulf, IndoMMLU \cite{koto2023large} incorporates Indonesia's local languages and cultures. Others include, CMMLU \cite{li2023cmmlu}, KMMLU \cite{son2024kmmlu},TurkishMMLU \cite{yuksel2024turkishmmlu}, PersianMMLU \cite{ghahroodi2024khayyam}, Global MMLU \cite{singh2024global} evaluates cultural and linguistic bias in 42 languages.

Despite these advancements, significant gaps remain for African LRLs and some Eastern European languages. Our work narrows this gap, by including Giriama, a LRL language spoken by approximately 2.6 million people in Kenya, which is currently absent from existing NLP benchmarks. Similarly, even though the Latvian language has been benchmarked before EU20-MMLU \cite{thellmann2024towards}, with relatively high-quality datasets, the datasets lack human-curated translations. As a result, they encounter challenges, including inaccurate auto-translations that compromise reliable evaluations. Giriama, in particular, has a limited digital presence and has historically been underrepresented in linguistic resources, making this the first multilingual benchmark developed explicitly for the language. Our contributions can be summarized as follows: 

\begin{itemize}
    \item \textbf{Novel Datasets:} we present LAG-MMLU, a curated dataset that includes Giriama's first gold-standard (human-translated) collection, alongside two silver-standard datasets for Latvian: one machine-translated and another machine-translated with subsequent human curation. Each dataset comprises 500 samples, randomly selected from the original MMLU benchmark. And will be publicly available for further research in multilingual NLP
    \item \textbf{Multilingual Comparative Analysis:}  We evaluate the performance of LLMs on two multilingual datasets, Giriama and Latvian, highlighting their understanding of complex knowledge representation.
    \item \textbf{Model Evaluation:} We evaluate the performance of eight (8) state-of-the-art (SOTA) closed and open-source LLMs using the Inspect AI framework. Our analysis includes both zero-shot and few-shot evaluations, with and without applying chain-of-thought (CoT) reasoning, for comprehensive insights.
    \item \textbf{Statistical Validation:} We employed confidence intervals (CI) and standard errors (SE) for statistical significance to ensure the robustness of our findings.
\end{itemize}

Driven by these considerations, the experiments in this paper aim to address the following key questions:
 
 \begin{itemize}
    \item \textbf{Q1:}  Which LLM exhibits the highest performance and comprehension across 55 tasks in both Latvian and Giriama?
    \item \textbf{Q2:} How does the performances of the three language datasets compare when evaluated across SOTA LLMs
    % \item \textbf{Q3:} Does human post-edited datasets improve benchmark quality compared to purely machine-translated datasets? 
    \item \textbf{Q3:}Does  chain-of-thought (CoT) reasoning enhance performance?
\end{itemize}

% In our work, we utilize the Massive Multitask Language Understanding (MMLU) benchmark, which covers 57 subjects ranging from STEM to humanities and social sciences. Our goal is to enhance the understanding of LLMs performance in low-resource languages, with a specific focus on Latvian and Giriama, and to contribute to the development of AI systems that are both linguistically and culturally inclusive.

\section{Related works}
\subsection{Multilingual Models Across Cultures}

The evaluation of the multilingual capabilities of LLMs has
garnered significant attention. This has led to increased research exploring their performance
across diverse linguistic landscapes. SOTA massively multilingual language Models (MMLMs) such as mBERT \cite{devlin2018bert}, XLMR \cite{conneau2019unsupervised}, and mT5 \cite{xue2020mt5} support 100+ languages worldwide and have shown exceptional proficiency in understanding text across diverse linguistic contexts. While encoder-decoder architectures like mT5 \cite{xue2020mt5} also demonstrate strong capabilities in natural language generation(NLG).

Additionally, generative models like GPT-4 \cite{achiam2023gpt}, LLaMA \cite{touvron2023llama}, BLOOM \cite{le2023bloom}, and BLOOMZ \cite{muennighoff2022crosslingual} are also gaining world recognition for their contributions to advancing natural language generation (NLG) and understanding. Significant challenges remain in ensuring cultural sensitivity and language equity \cite{dawson2024evaluating}. 

Several research efforts have shown that multilingual models perform well in high-resource languages (HRLs) like English and French, but still struggle with very  LRLs \cite{li2024quantifying,hedderich2020survey,ranathunga2022some}, particularly in Africa \cite{adelani2021masakhaner,alabi2022adapting, adebara2024cheetah} and South Asia \cite{lahoti2022survey, baruah2021low}, due to limited training data \cite{adebara2024cheetah, magueresse2020low}. Challenges such as cultural nuances \cite{romero2024cvqa, winata2024worldcuisines}, dialectal variation \cite{faisal2024dialectbench}, and code-switching \cite{winata2021multilingual, etori2024rideke} further hinder model performance. While efforts like cross-lingual transfer learning and culturally relevant datasets have been made to address these issues \cite{hu2020xtreme, winata2022cross,liu2021importance}, performance gaps persist in underrepresented languages, especially African \cite{adelani2024irokobench}.

\subsection{Multilingual LLMs  Evaluation }
The evaluation of LLMs has evolved significantly with the development of various benchmarks to assess their performance across tasks and languages. Early benchmarks like GLUE \cite{wang2018glue} and SuperGLUE \cite{wang2019superglue} focused on natural language understanding (NLU) tasks in English, using datasets of different sizes. Multilingual evaluation later gained attention with benchmarks such as XGLUE \cite{liang2020xglue} XTREME \cite{hu2020xtreme}, and XTREME-R \cite{ruder2021xtreme}, which extended the evaluation to over 20 languages, emphasizing cross-lingual transfer capabilities, however, the coverage of Baltic and African languages in XTREME is limited. For natural language generation (NLG), the GEM benchmark introduced by  \newcite{gehrmann2021gem} includes tasks like translation, summarization, and description generation in multiple languages.

As LLMs became extensive and with near human-level performance, evaluation tasks shifted to emphasize reasoning and real-world knowledge. Benchmarks like MMLU \cite{hendrycks2020measuring},  HellaSwag \cite{zellers2019hellaswag}, HELM \cite{liang2022holistic}, LAMBADA \cite{paperno2016lambada} and WinoGrande \cite{sakaguchi2021winogrande} became standards for assessing reasoning abilities and knowledge representation. Models like GPT-4o and LLaMA have been evaluated on some benchmarks, with GPT-4o achieving human-level performance on various tasks \cite{hurst2024gpt}.

% Various benchmarks have been released to evaluate pre-trained LMsMost existing multilingual NLP benchmarks such as \cite{hendrycks2020measuring, hu2020xtreme, wang2018glue, wang2019superglue, guzman2019flores} are heavily skewed toward high-resource languages, particularly those in the Indo-European language family, and reflect predominantly Western cultural contexts. As a result, these benchmarks fail to capture the linguistic and cultural diversity of the global population, making them less reliable in assessing the performance of multilingual language models (MMLMs) across underrepresented languages and cultures \cite{bender2019benderrule}. 

Recent research has focused on the development of multilingual datasets to more accurately capture linguistic and cultural diversity. Initiatives such as CVQA \cite{romero2024cvqa}, WorldCuisines \cite{winata2024worldcuisines}, D-PLACE \cite{kirby2016d}, Wikipedia Cultural Diversity Dataset \cite{miquel2019wikipedia}, TeDDi Sample \cite{moran2022teddi}, Cheetah \cite{adebara2024cheetah}, Irokobench \cite{ifeoluwa2024irokobench}, and CulturalBench \cite{chiu2024culturalbench}  have made significant progress in improving the representation within multilingual models. 

Leveraging community-driven approaches to develop localized datasets ensures that diverse linguistic and cultural nuances are effectively incorporated into language technologies. Underscoring cultural context in which language is used rather than relying solely on translation-based approaches \cite{tiedemann2020tatoeba}.

\subsection{Human evaluation of multilingual models}
The development of evaluation datasets for specific languages involves significant human resources.
Therefore, a widely used strategy is to apply machine translation, with or without manual post-editing \cite{dargis2024evaluating}.

Human ability to understand language is general, flexible, and robust \cite{wang2018glue, lin-och-2004-orange}. Hence, human evaluations are typically considered the gold standard in NLG to assess the effectiveness of multilingual models \cite{clark2021all, chiang2023can}, particularly in evaluating their ability to generate text that aligns with diverse linguistic and cultural contexts. 

The use of automatic metrics such as BLEU \cite{Papineni02bleu:a} and ROUGE \cite{lin-2004-rouge}, even though commonly used, often fail to capture cultural nuances, making human evaluation essential for a more comprehensive assessment \cite{kocmi2021ship}. Human evaluations are essential for assessing how well multilingual models handle grammatical, syntactical, and contextual differences, particularly in LRLs  \cite{costa2022no}. Human raters are better at identifying these nuances, using criteria such as appropriateness, bias detection, and cultural sensitivity \cite{choenni2024evaluation}.

\begin{table*}[ht]
\small
\centering
\begin{tabular}{p{3.2cm}|p{11.8cm}} % Adjust column widths here
\hline
\textbf{Language} & \textbf{Question and Answers (Question subject: miscellaneous)} \\ \hline

\textbf{English} & According to the children's nursery rhyme what type of ocean did Columbus sail in 1492? \\
 & A: calm \textcolor{red}{\textbf{X}}, B: blue \textcolor{green}{\textbf{\checkmark}}, C: windy \textcolor{red}{\textbf{X}}, D: really big \textcolor{red}{\textbf{X}} \\ \hline

\textbf{Giriama} & Kulingana na wira wa kitalu cha ahoho ni aina yani ya bahari ambayo Columbus wasafiri makathi ga 1492? \\
 & A: Kuhurira \textcolor{red}{\textbf{X}}, B: buluu \textcolor{green}{\textbf{\checkmark}}, C: peho \textcolor{red}{\textbf{X}}, D: bomu jeri \textcolor{red}{\textbf{X}} \\ \hline

\textbf{Latvian } & Saskaņā ar bērnu bērnudārza atskaņu, kāda veida okeānu Kolumbs kuģoja 1492. gadā?\\
 (autotranslated) & A: Mierīgs \textcolor{red}{\textbf{X}}, B: zils \textcolor{green}{\textbf{\checkmark}}, C: Vējains \textcolor{red}{\textbf{X}}, D: Ļoti liels \textcolor{red}{\textbf{X}} \\ \hline

 \textbf{Latvian} & Saskaņā ar bērnudārza pantiņu, kāda veida okeānu Kolumbs kuģoja 1492. gadā?\\
 (autotranslated \& edited)& A: Mierīgu \textcolor{red}{\textbf{X}}, B: Zilu \textcolor{green}{\textbf{\checkmark}}, C: Vējainu \textcolor{red}{\textbf{X}}, D: Ļoti lielu \textcolor{red}{\textbf{X}} \\ \hline

\end{tabular}
\caption{\footnotesize \textbf{Sample question translated into Giriama and Latvian (AT: autotranslated, AT+E: autotranslated and edited) with correct answers marked \textcolor{green}{\textbf{(\checkmark)}} and incorrect answers marked \textcolor{red}{\textbf{(X)}}. The correct answer "blue" in English refers to the popular children's rhyme "In 1492, Columbus sailed the ocean blue," which is a cultural reference that may not resonate in Latvian or Giriama without further explanation.}}
\label{tab:translations}
\end{table*}

\section{Methodology}
\subsection{LAG-MMLU Dataset}
 We carefully curated a shuffled subset of 500 samples from the original MMLU benchmark \cite{hendrycks2021measuringmassivemultitasklanguage}. The original MMLU contains over 15,000 multiple-choice questions (MCQs) spanning 57 subjects across disciplines such as social science, history, and STEM. Sourced from publicly available materials, including practice exams like the GRE and USMLE. The dataset categorizes questions from elementary to advanced professional levels.

\subsection{Dataset Selection Process}
To ensure a representative and diverse evaluation dataset, we began by shuffling the full MMLU dataset. The shuffling step ensured randomization across topics, avoiding biases from subject-specific clustering. Following this, the top 500 entries were selected for inclusion. The dataset was then formatted using a custom pipeline that maps records to standardized MCQ samples, preserving key fields such as questions, answers, and metadata. To ensure an evenly distributed, high-quality subset. 

\subsection{Languages covered}
LAG-MMLU evaluates two languages: 

\paragraph{Latvian (lav):} Latvian\footnote{\cite{smith2004latvia}} is an East Baltic language that belongs to the Indo-European language family and is spoken in the Baltic region. With around 2 million speakers worldwide, Latvian is the official language of Latvia and is spoken as a second language by over 500,000 non-Latvians within the country \cite{smith2004latvia}. Latvian is more morphologically complex \cite{paikens2024computational} than English, and there is still limited research and representation in widely used multilingual benchmarks on NLP for the language. The complexity further adds to the difficulty of LLMs processing \cite{dargis2024balsutalka}.

\paragraph{Giriama (nyf):} 
Giriama (Kigiryama)\footnote{\href{https://en.wikipedia.org/wiki/Giriama_people}{Giriama people}}, is a Bantu language spoken by approx. 700,000 people, primarily in the coastal part of Kenya. It is one of the nine (9) ethnic groups or tribes that make up Mijikenda languages—classified under the Northeastern Bantu subgroup of the Niger-Congo family. Predominantly oral, Giriama has limited digital texts, though recent efforts have promoted literacy using the Latin alphabet. Despite these efforts, Giriama remains under-resourced regarding linguistic and digital resources.

\subsubsection{Dataset collection}
\begin{figure*}[ht]
    \centering
    \includegraphics[width=0.95\textwidth]{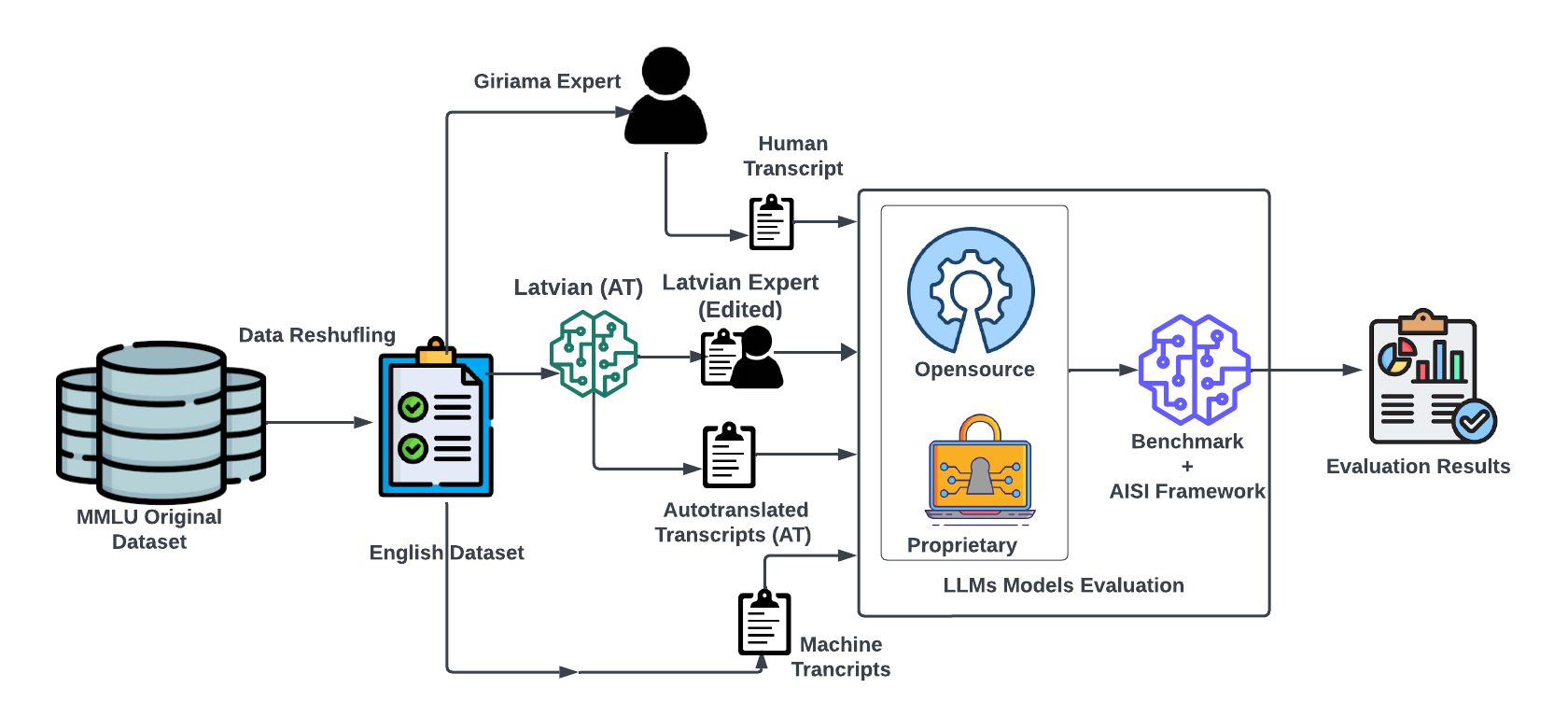}
    \caption{\textbf{Frontier LLMs in Latvian and Giriama Dataset and Benchmarking pipeline}}
    \label{fig:ccom}
\end{figure*}
Our random shuffled dataset collected in three sets:

\paragraph{Baseline Dataset:} The process began with the original MMLU English dataset, comprising 500 questions. These questions form the baseline, for comparison and evaluation in subsequent steps.

\paragraph{Machine Translations:} The baseline questions were translated into Latvian using the MyMemory API \footnote{https://mymemory.translated.net/}\cite{mymemory_translation_memory}. The MyMemory API is a machine translation (MT) system that combines translation memory (TM) with statistical methods. Using a large repository of human-translated texts retrieves relevant segments for accurate, context-aware translations, hence a hybrid approach.

\paragraph{Human-Edited Translations:} To ensure accuracy and cultural inclusivity, the machine-translated Latvian questions were reviewed and curated by human translators. Additionally, human experts who are native speakers, translated the English questions into Giriama, to ensure high-quality datasets.

\subsubsection{Dataset Statistics}
The dataset comprises 55 tasks across four domains: \textbf{Humanities} (34\%, 170 tasks), \textbf{STEM} (22\%, 112 tasks), \textbf{Social Sciences} (20\%, 99 tasks), and \textbf{Other} (24\%, 119 tasks). Professional law (57 tasks), moral scenarios (31 tasks), mathematics (17 tasks), psychology (47 tasks across subfields), and clinical knowledge (15 tasks) have the highest representations. The random selection effectively captured 55 of the original 57 tasks, hence 96.5\% retention rate as shown in Table~\ref{tab:subject_distribution} at the Appendix.

\begin{figure}[h]
    \centering
    \includegraphics[width=\linewidth]{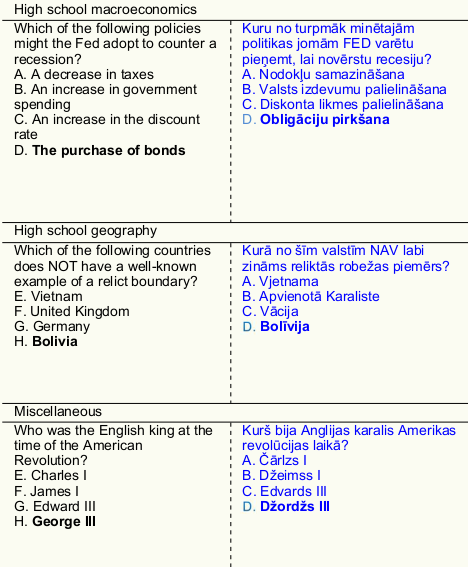}
    \caption{Examples of English-Latvian translated; topics include high-school macroeconomics, high-school geography, and miscellaneous}
    \label{fig:translated_questions}
\end{figure}

\begin{figure}[h]
    \centering
    \includegraphics[width=\linewidth]{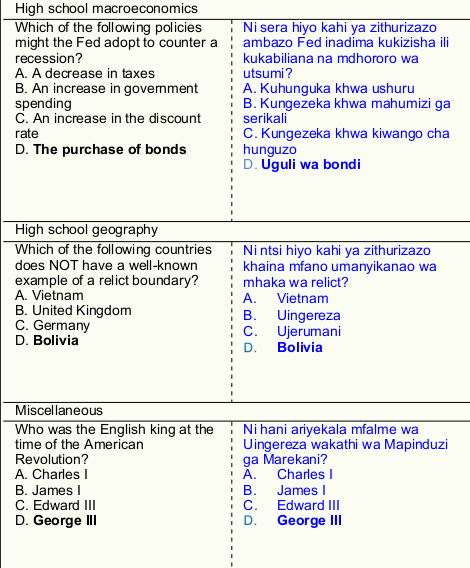}
    \caption{Examples of English-Giriama translated; topics such as high-school macroeconomics, high-school geography, and miscellaneous}
    \label{fig:translated_questions2}
\end{figure}

\subsubsection{Translations and annotation }
We recruited one Giriama language coordinator, who also doubled as a translator. Holds a master’s degree in computer science and is a native speaker of the language with extensive experience as a translator. As for the Latvian language,  a Latvian-Native translator and annotator reviewed and edited the machine-translated questions. The goal is to correct any errors that could hinder comprehension or lead to misinterpretations of the answers.

\textbf{Quality Check:} Language coordinators thoroughly reviewed and corrected poorly translated sentences to preserve the highest quality. The translator was compensated after the quality assurance phase, aligning payments with the local living wage to recognize their efforts fairly.

\subsection{Task Overview}
We evaluate the multilingual understanding of open and closed LLMs using translated questions from the MMLU benchmark, which tests world knowledge and problem-solving skills in zero-shot and few-shot settings. We assess model performance across three languages—English, Latvian, and Giriama—using human-annotated translations of 500 questions. 
\subsection{Inspect Evaluation Framework}
We employed the UK AISI Inspect framework \cite{inspect2024} \footnote{\url{https://github.com/UKGovernmentBEIS/inspect_ai}} to evaluate LAG-MMLU benchmark across zero-shot, few-shot, and CoT reasoning. Inspect facilitates modular evaluation through dynamic prompt engineering, dataset filtering, and scoring metrics such as exact match and choice-based evaluation. 

\subsection{Model Choice}
We used closed and open large LLMs to evaluate their performance. \footnote{Models: claude-3-5-sonnet-20241022, gemini-1.5-pro-002, gpt-4o-2024-08-06, Meta-Llama-3.3-70B-Instruct-Turbo, mistral-large-2407,o1, DeepSeek-V3 and DeepSeek-R1} As shown in Table~\ref{tab:main_results_table}.

\textbf{Closed LLMs:}
The four closed SOTA models selected for this study include OpenAI o1 \cite{jaech2024openai}, Claude 3.5 Sonnet \cite{claude-3.5-sonnet-2024}, GPT-4o \cite{hurst2024gpt}, and Gemini 1.5 pro \cite{team2024gemini}. These models were chosen due to their established performance and popularity in handling
various tasks. Recent work has shown that proprietary models exhibit better multilingual capabilities \cite{ahuja2023megaverse}. However, their pre-training, fine-tuning processes and model size are not disclosed.

\textbf{Open LLMs:} We evaluate four open models: Meta-Llama-3.3 70B IT \cite{dubey2024llama},  Mistral-Large \cite{jiang2023mistral},DeepSeek-V3 \cite{liu2024deepseek} and DeepSeek-R1 \cite{guo2025deepseek}. Both decoder-only models are effective for multilingual and LRLs. These models are openly available under licenses ranging from research to fully open and permissive, providing transparency in their architecture. However, the exact composition of their training data is, in fact, not known. Determining whether Giriama or Latvian is included in the pretraining data of models is challenging.

\subsection{Evaluation metrics}
We evaluated the performance of all LLMs with an accuracy score. For each model, accuracy was computed as the proportion of correct outputs from a test set comprising 500 samples. We calculate the (binomial) standard error to account for uncertainty in the performance estimates. 

\textbf{0-Shot Evaluation:}
We implement the 0-shot evaluation, on the entire dataset without filtering based on the number of available examples per subject. In this configuration, the prompt provided to the model consists solely of the test question.

\textbf{Few-Shot Evaluation:}
We employ a 5-shot prompt-based evaluation, retaining subjects with at least six samples—one reserved as the test instance. For each test sample, the function \texttt{mmlu\_n\_shot\_solver(n=5)}retrieves five example question-answer pairs from the subject's development set. These are formatted using the \texttt{MMLU\_MULTISHOT\_PROMPT\_TEMPLATE}, which presents the examples followed by the test question. The model is then conditioned on this prompt, generating a single-token answer via \texttt{generate(max\_tokens=1)}.See Figure~\ref{fig:mmlu_template}.

% we adopt a 5-shot prompt-based evaluation. First, we filter the dataset to retain only those subjects for which at least six samples are available. One sample is reserved as the test instance. For each test sample, a custom solver—implemented via the function \texttt{mmlu\_n\_shot\_solver(n=5)}-retrieves five example question-answer pairs from the corresponding subject's development set. These examples are then formatted into a cohesive pre-defined template \texttt{MMLU\_MULTISHOT\_PROMPT\_TEMPLATE}, which presents the five examples followed by the test question. The model is subsequently conditioned on this multi-shot prompt, and a single-token answer is generated (through \texttt{generate(max\_tokens=1)} function). 

\begin{table*}[t]
\centering
\small
\setlength{\tabcolsep}{8pt}  % Increase spacing between columns
\renewcommand{\arraystretch}{1.4}  % Increase row height
\begin{adjustbox}{max width=\textwidth}  % Ensures the table fits without shrinking text
\begin{tabular}{llcccccc}
\toprule
\textbf{Model} & \textbf{COT}  
  & \multicolumn{2}{c}{\textbf{English Accuracy (\%)}} 
  & \multicolumn{2}{c}{\textbf{Latvian Accuracy (\%)}} 
  & \multicolumn{2}{c}{\textbf{Giriama Accuracy (\%)}} \\
\cmidrule(lr){3-4}\cmidrule(lr){5-6}\cmidrule(lr){7-8}
&  
  & \textbf{0-shot} & \textbf{5-shot}  
  & \textbf{0-shot} & \textbf{5-shot}  
  & \textbf{0-shot} & \textbf{5-shot}  \\
\midrule
\textbf{Open AI o1} & {\color{red}\ding{55}}  
    & \textbf{92.8} [90.5, 95.1] & -- 
    & \textbf{88.8} [86.0, 91.6] & -- 
    & \textbf{70.8} [66.8, 74.8] & -- \\
\midrule
\multirow{2}{*}{GPT-4o} 
  & {\color{red}\ding{55}}  & 83.6 [80.4, 86.8] & --  
       & 75.6 [71.8, 79.4] & 78.0 [74.4, 81.6]  
       & 53.2 [48.8, 57.6] & --  \\
  & {\color{green}\ding{51}}  & -- & 84.5 [81.3, 87.7]  
       & 79.5 [76.0, 83.0] & --  
       & -- & 62.3 [58.1, 66.5]  \\
\midrule
\multirow{2}{*}{Claude 3.5}  
  & {\color{red}\ding{55}}  & 88.8 [86.0, 91.6] & --  
       & 78.2 [74.6, 81.8] & --  
       & 54.8 [50.4, 59.2] & --  \\
  & {\color{green}\ding{51}}  & -- & 87.1 [84.2, 90.0]  
       & -- & 83.0 [79.7, 86.3]  
       & -- & 60.9 [56.6, 65.2]  \\
\midrule
\multirow{2}{*}{Gemini 1.5 Pro}  
  & {\color{red}\ding{55}}  & 84.0 [80.8, 87.2] & --  
       & 70.3 [66.3, 74.3] & --  
       & 51.4 [47.0, 55.8] & --  \\
  & {\color{green}\ding{51}}  & -- & 85.4 [82.3, 88.5]  
       & -- & 79.9 [76.4, 83.4]  
       & -- & 61.8 [57.5, 66.1]  \\
\midrule
\multirow{2}{*}{DeepSeek-V3}  
  & {\color{red}\ding{55}}  & 87.6 [84.7, 90.5] & --  
       & 70.7 [66.7, 74.7] & --  
       & 49.8 [45.4, 54.2] & --  \\
  & {\color{green}\ding{51}}  & -- & 87.9 [85.0, 90.8]  
       & -- & 76.1 [72.4, 79.8]  
       & -- & 52.3 [47.9, 56.7]  \\
\midrule
Llama-70B IT  & {\color{red}\ding{55}}  
    & 81.0 [77.6, 84.4] & --  
    & 57.3 [53.0, 61.6] & --  
    & 41.0 [36.7, 45.3] & --  \\
\midrule
Mistral-large  & {\color{red}\ding{55}}  
    & 80.0 [76.5, 83.5] & --  
    & 49.3 [44.9, 53.7] & --  
    & 35.6 [31.4, 39.8] & --  \\
DeepSeek-R1  & {\color{red}\ding{55}}  
    & 81.4 [78.0, 84.8] & --  
    & 70.2 [66.2, 74.2] & --  
    & 52.2 [47.8, 56.6] & --  \\
\bottomrule
\end{tabular}
\end{adjustbox}
\caption{\textbf{Performance of Models on LAG-MMLU.} The table shows Accuracy (\%) with 95\% Confidence Intervals (CI) across three languages (English, Latvian, and Giriama) under 0-shot and 5-shot settings. Results are presented for models with and without Chain-of-Thought (CoT) prompting, where ({\color{red}\ding{55}}) represents No CoT and ({\color{green}\ding{51}}) represents CoT. (Also See Table~\ref{tab:main_results_table2}). All experiments were conducted with the temperature \textbf{set to 0}. \textit{OpenAI-o1} achieves the highest overall accuracy (\textbf{92.8\%}) across the languages. The symbol \texttt{--} indicates that no experiment was conducted for that setting.}
\label{tab:main_results_table}
\end{table*}

\begin{figure*}[h]
    \centering
    \includegraphics[width=\textwidth]{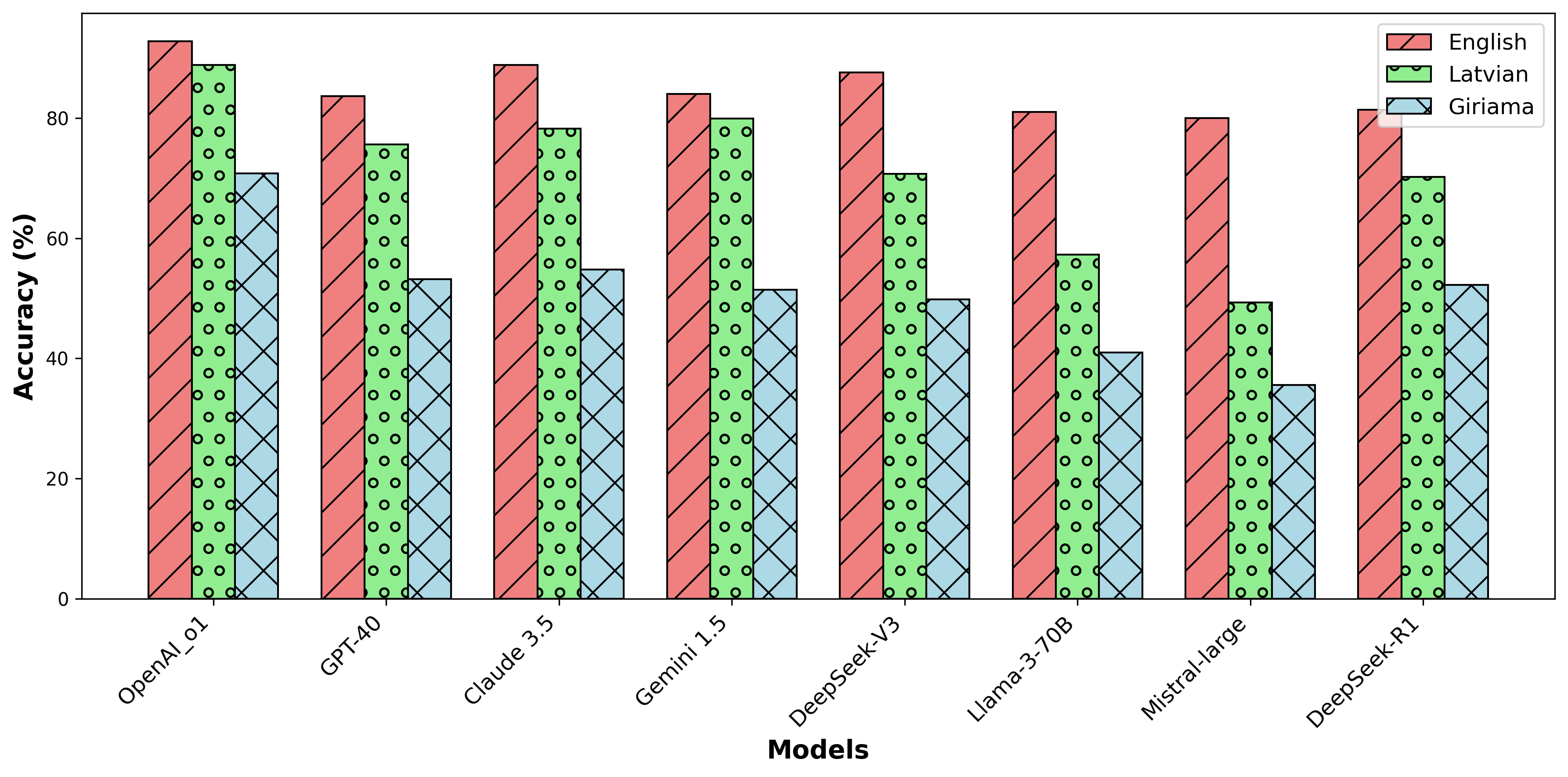} % Ensure filename matches exactly
    \caption{Performance of All Models on LAG-MMLU on English, Latvian, and Giriama (0-shot). OpenAI-o1 shows the best performance then claude 3.5 Sonnet. Mistral-large underperforms in LRLs.}
    \label{fig:lag_mmlu_perf}
\end{figure*}

\begin{figure}[t]
    \centering
    \includegraphics[width=0.95\columnwidth]{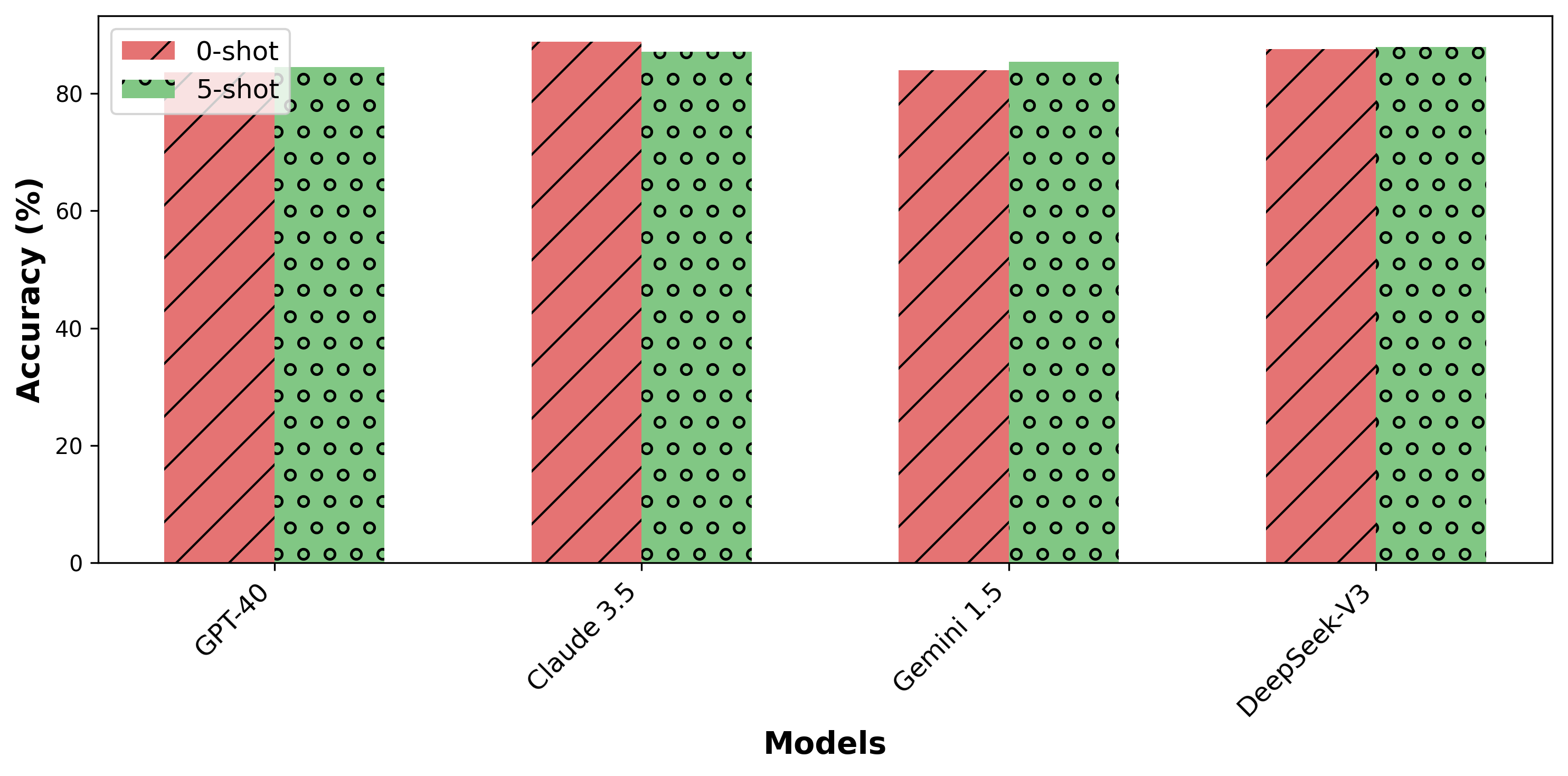}
    \caption{0-shot vs. 5-shot performance across four models. Few-shot learning improves most models, except Claude 3.5, which drops by 1.7\%.}
    \label{fig:o_shot_vs_5_shot}
\end{figure}

\begin{table*}[t]
\centering
\footnotesize  % or \small
\setlength{\tabcolsep}{4pt}
\renewcommand{\arraystretch}{1.1}
\resizebox{\textwidth}{!}{%
\begin{tabular}{llcccccc}
\toprule
 &  & \multicolumn{2}{c}{\textbf{English}} & \multicolumn{2}{c}{\textbf{Latvian}} & \multicolumn{2}{c}{\textbf{Giriama}} \\
\cmidrule(lr){3-4}\cmidrule(lr){5-6}\cmidrule(lr){7-8}
\textbf{Model} & \textbf{COT} 
  & \textbf{0-shot} & \textbf{5-shot} 
  & \textbf{0-shot} & \textbf{5-shot} 
  & \textbf{0-shot} & \textbf{5-shot} \\
\midrule
\multirow{2}{*}{GPT\_4o} 
  & \xmark  & 0.830 $\pm$ 0.017 & -- 
  & 0.762 $\pm$ 0.019 & 0.790 $\pm$ 0.018 
  & 0.514 $\pm$ 0.022 & -- \\
  & \cmark & -- & 0.847 $\pm$ 0.016 
  & -- & 0.802 $\pm$ 0.018 
  & -- & 0.582 $\pm$ 0.022 \\
\bottomrule
\end{tabular}
}% end resizebox
\caption{\textbf{Performance of GPT\_4o.} Accuracy has standard error (±) under 0-shot and 5-shot settings for English, Latvian, and Giriama, using the temperature \textbf{set =1}. Results are shown for both the \textbf{no CoT} (\xmark) and \textbf{CoT} (\cmark). The symbol \texttt{--} indicates that no data is reported.}
\label{tab:GPT4o_results}
\end{table*}

\section{Experiment Setup}
We evaluate model performance using:

\paragraph{\textbf{Shots, CoT, and Temperature Settings:}} We evaluate our Models under 0-shot and 5-shot settings, with and without chain-of-thought (CoT) reasoning. A temperature of 0 was applied uniformly for deterministic behavior, with GPT-4o also tested at temperature 1 (Table~\ref{tab:GPT4o_results}). The evaluation covered three datasets, comprising 55 tasks across four categories (Figure~\ref{fig:lag_mmlu_distribution}). \footnote{\url{https://github.com/akanepajs/capabilities_lv_giriama/tree/main/MMLU}}

\subsection{Results and Discussion}
\subsubsection{Results across all models}
Table \ref{tab:main_results_table} presents the performance of all LLMs evaluated in three languages—English, Latvian, and Giriama—in both zero-shot and five-shot settings, with and without CoT prompting. Performance is reported as accuracy with confidence intervals (CIs). As expected, the \textbf{OpenAI-o1} model stands out as the top performer among both closed and open models, with remarkable accuracy of \textbf{92.8\%} in English, \textbf{88.8\%} in Latvian, and \textbf {70.8\%} in the 0-shot test without chain-of-thought (CoT), reinforcing its general ability to handle a broad spectrum of tasks effectively in multiple languages and serves as the benchmark for other models. Even in LRLs like Latvian or Giriama, they tend to maintain a comfortable margin over other models. In addition, \textbf{Claude 3.5 Sonnet} demonstrates strong results, achieving \textbf{87.1\%} in 5-shot, further solidifying its position as a solid closed model contender. \textbf{Gemini-1.5-pro} also shows competitive performance, reaching \textbf{85.4\%} in the 5-shot test, but still trails behind its closed model's counterparts, particularly the GPT-4o.

The open-sourced models show some impressive performances, albeit generally weaker than closed-source. \textbf{DeepSeek-V3} leads the open models, achieving {87.6\%} in the 0-shot test and \textbf{87.9\%} in the 5-shot test. While these results indicate strong performance, particularly in the 5-shot task, they still fall short of matching the best-performing closed models across languages. For instance, the open models like \textbf{Mistral-Large}, \textbf{Llama 3.3 70B IT},  have low accuracies across the tasks, especially on Latvian and Giriama with a CI of \textbf{31.4\%-45.3\%}.

The results reinforce the strengths of \textbf{closed-source models}in providing a robust performance. This advantage is apparent in multilingual settings (across resource-rich and lower-resource languages like Latvian or Giriama) and in their ability to approach complex tasks with minimal prompting. Nevertheless, open models like \textbf{DeepSeek-V3} still manage to showcase competitive performance, particularly when prompted with 5-shot examples, open models hold potential for unique use cases where customization, transparency, or self-hosting becomes critical factors. This performance gap also underscores that, while the top closed models set a high benchmark, open models remain valuable contenders.

Llama-70B IT and Mistral-Large both exhibit weaker performance, scoring around 80.0–81.0\% under the 0-shot English. while DeepSeek-R1 achieves a slightly higher accuracy of 81.4\%. However, their performance drops significantly in LRLs, with Llama-70B IT 57.3\% in Latvian and 41.0\% in Giriama, and Mistral-Large performing even lower at 49.3\% (Latvian) and 35.6\% (Giriama). In contrast, DeepSeek-R1 maintains a relatively stronger performance at 70.2\% (Latvian) and 52.2\% (Giriama). Still, none of them reach the higher English accuracy of DeepSeek-V3 (87.6\% in 0-shot).

\subsubsection{Performance Across All Languages}
Table~\ref{tab:main_results_table} demonstrates that across all conditions all evaluated models perform best in English, which is unsurprising given the abundant training data available for this language, followed by Latvian, with Giriama consistently showing the lowest scores. For instance, the closed‑source OpenAI-o1 model reaches an accuracy of 92.8\% in English, while its performance in Latvian and Giriama—though still high—shows a measurable decline (with Latvian accuracy at 88.8\% and Giriama dropping further to 70.8\%).  

In contrast, open‑source models such as \textbf{Llama3.3 70B IT} and \textbf{Mistral‑Large} exhibit a weak performance drop when shifting from English to these LRLs. For example, while \textbf{DeepSeek‑V3} scores competitively in English (87.6\% in the 0‑shot setting), its accuracy in  Giriama falls by a larger margin relative to the closed‑source models. closed‑source models maintain a relatively stable performance across all languages, likely due to extensive pretraining with robust multilingual data and continual fine‑tuning. Despite their black boxes; their internal architectures, training data, and fine‑tuning methodologies remain largely opaque.

\subsubsection{CoT vs. non-CoT Performance}
Many benchmark results such as  CMMLU \cite{li2023cmmlu}, GSM8K\cite{zhong2024achieving}, and KMMLU \cite{son2024kmmlu} have demonstrated that CoT prompting has improved the performance on multitask evaluations.
Table~\ref{tab:main_results_table} and Figure~\ref{fig:lag_mmlu_perf} demonstrate that enabling CoT prompting yields substantial accuracy improvements for most models. We have four models---GPT{4\_o}, Claude 3.5 Sonnet, Gemini 1.5 Pro, and DeepSeek-V3--- that employ CoT prompting, and they generally achieve significant accuracy gains compared to their no-CoT baselines. For example, Claude 3.5 Sonnet sees an increase from 54.8\% to 60.9\% in the Giriama 0-shot setting when CoT is enabled, but there is a slight decrease in performance on the English language when CoT is enabled (from 88.8\% to 87.1\%). GPT{4\_o} experiences a comparable improvement from 75.6\% to 79.5\% in the Latvian 0-shot setting when CoT is enabled. This pattern suggests that both closed-source and open-sourced, benefit most from CoT’s structured approach. COT has minimal effect on English performance, as models already achieve high zero-shot accuracy (e.g., GPT-4o: 0.836 → 0.845).

\subsubsection{ Few-Shot Performance}
Shifting from 0‑shot to 5‑shot prompts benefits every model in Table~\ref{tab:main_results_table} and Table~\ref{tab:GPT4o_results}, both closed-source and open-source models benefit from few-shot prompting; however, the degree of improvement from 0-shot to 5-shot vary across models. In our experiments, the 5-shot setting—proved to be a critical factor in enhancing model performance. GPT-4o shows a 0.9\% improvement, indicating a moderate benefit from few-shot learning. Gemini 1.5 gains the most, with a 1.4\% increase, while DeepSeek-V3 experiences a smaller improvement of 0.3\%. However, Claude 3.5 Sonnet performs worse with few-shot learning, exhibiting a 1.7\% decrease in accuracy as shown in Figure~\ref{fig:o_shot_vs_5_shot}.

\section{Conclusion}
We introduce the LAG-MMLU Benchmark, a multilingual evaluation dataset comprising 500 MCQs across 55 subjects or tasks, originally sourced from MMLU. The dataset includes: (1) human-translated Giriama questions and (2) machine-translated, human-edited Latvian questions. This benchmark is designed to assess multi-task language understanding capabilities in Latvian and Giriama.
Our findings reveal significant performance gaps in existing SOTA LLMs, particularly in Giriama, where models underperformed even in few-shot settings. \textbf{Gemini 1.5 Pro} exhibits the most significant improvement with CoT enabled, gaining \textbf{+9.6} and \textbf{+10.4} respectively in \textbf{Latvian} and \textbf{Giriama. }English performance remains strong, \textbf{Latvian} and \textbf{Giriama} results indicate substantial room for improvement, reinforcing the need for dedicated training data, fine-tuning, and linguistic adaptations for LRLs. We believe that the LAG-MMLU Benchmark will serve as a valuable resource for the LRL research community.

\section{Limitations}

LAG-MMLU benchmark offers valuable insights for the NLP community but faces several limitations that future work must address. As mentioned, native speakers curated the dataset to ensure high translation quality, but errors from automatic translation persist. For instance, 6\% of the MMLU dataset \cite{miller2024adding}included mistranslations such as "Not wrong, wrong" rendered as "Not wrong, Not wrong" in Latvian, increasing the English-Latvian error gap by 2\%. 

Some culturally specific questions remain challenging to translate accurately. The small sample size limits generalizability, and lack formal quality control measures, such as inter-annotator agreement (IAA) or computational frameworks like AfriCOMET \cite{wang2023afrimte}, further impacts the validity of results. Statistical techniques such as clustered standard error calculations \cite{miller2024adding}) could enhance analysis reliability.

The black-box nature of closed models introduces uncertainty and undermines fair comparisons. LAG-MMLU represents only a fraction of the world’s many languages \cite{singh2024global}, hence neglecting many dialects, creoles\cite{robinson2024krey}, code-switching \cite{ etori2024rideke}. Future work should include more languages and geo-cultural to improve inclusivity.

\section{Ethical considerations}

The MMLU dataset includes highly technical terms from fields such as law, science, and ethics (e.g., "neurotransmitters" and "Pauli exclusion principle"), which are difficult to translate into LRLs like Giriama as shown in Table~\ref{tab:translations}. A lack of established vocabulary or the need for highly specialized translators often leads to inconsistent or unclear translations. Additionally, the dataset reflects Western-centric knowledge and perspectives, which can misrepresent cultural nuances such as political ideologies or gender roles when translated. The linguistic and cultural challenges undermine language models' accuracy, consistency, and real-world applicability in the target culture.

\section{Acknowledgements}
We appreciate our annotators' exceptional effort and dedication to producing high-quality work.

\bibliographystyle{acl_natbib}
\bibliography{nodalida2025}

\newpage
\appendix
\onecolumn

\section{Few-Shots Templates} 
\label{sec:shot_ templates}

\begin{figure*}[h]
    \centering
    \includegraphics[width=.9\textwidth]{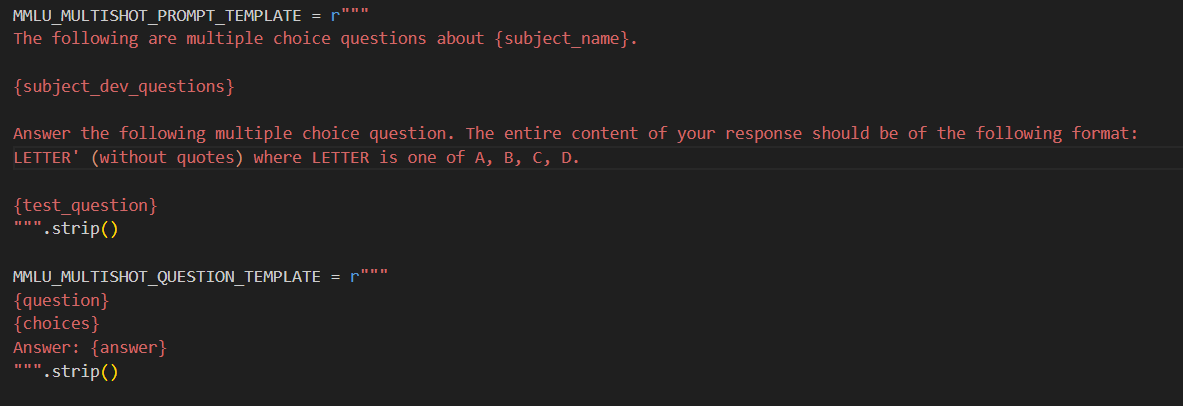}  % Ensure the filename matches your uploaded image
    \caption{MMLU Multi-Shot Prompt and Question Template. The templates define the format for multiple-choice question prompts and individual test questions, ensuring consistent structure across different subject areas.}
    \label{fig:mmlu_template}
\end{figure*}

\newpage

\section{Datasets Statistics} 
\label{sec:data_stats}

\begin{table*}[h]
\centering
\footnotesize
\renewcommand{\arraystretch}{1.1}
\setlength{\tabcolsep}{8pt} % Adjust spacing between columns
\begin{adjustbox}{max width=\textwidth}
\begin{tabular}{llc|llc}
\toprule
\multicolumn{3}{c|}{\textbf{Humanities}} & \multicolumn{3}{c}{\textbf{STEM}} \\
\midrule
\textbf{Subject} & \textbf{Freq} &  & \textbf{Subject} & \textbf{Freq} &  \\
\midrule
moral\_disputes & 12 & & abstract\_algebra & 3 &  \\
professional\_law & 57 & & anatomy & 1 &  \\
formal\_logic & 6 & & astronomy & 4 &  \\
high\_school\_european\_history & 3 & & college\_chemistry & 2 &  \\
high\_school\_us\_history & 9 & & college\_computer\_science & 2 &  \\
high\_school\_world\_history & 12 & & college\_mathematics & 5 &  \\
international\_law & 2 & & college\_physics & 4 &  \\
jurisprudence & 4 & & computer\_security & 4 &  \\
logical\_fallacies & 9 & & conceptual\_physics & 6 &  \\
moral\_scenarios & 31 & & electrical\_engineering & 7 &  \\
philosophy & 9 & & elementary\_mathematics & 17 &  \\
prehistory & 11 & & high\_school\_biology & 7 &  \\
world\_religions & 5 & & high\_school\_chemistry & 10 &  \\
\textbf{Total} & \textbf{170} & & high\_school\_computer\_science & 3 &  \\
 &  & & high\_school\_mathematics & 16 &  \\
 &  & & high\_school\_physics & 6 &  \\
 &  & & high\_school\_statistics & 10 &  \\
 &  & & machine\_learning & 5 &  \\
\textbf{} & \textbf{} & & \textbf{Total} & \textbf{112} &  \\
\cmidrule{1-3} \cmidrule{4-6}
\multicolumn{3}{c|}{\textbf{Social Sciences}} & \multicolumn{3}{c}{\textbf{Other}} \\
\midrule
\textbf{Subject} & \textbf{Freq} &  & \textbf{Subject} & \textbf{Freq} &  \\
\midrule
econometrics & 1 & & business\_ethics & 4 &  \\
high\_school\_geography & 9 & & clinical\_knowledge & 15 &  \\
high\_school\_government\_and\_politics & 4 & & college\_medicine & 3 &  \\
high\_school\_macroeconomics & 8 & & global\_facts & 5 &  \\
high\_school\_microeconomics & 5 & & human\_aging & 8 &  \\
high\_school\_psychology & 21 & & management & 4 &  \\
human\_sexuality & 7 & & marketing & 19 &  \\
professional\_psychology & 26 & & miscellaneous & 30 &  \\
public\_relations & 2 & & nutrition & 7 &  \\
security\_studies & 4 & & professional\_accounting & 9 &  \\
sociology & 10 & & professional\_medicine & 8 &  \\
us\_foreign\_policy & 2 & & virology & 7 &  \\
\textbf{Total} & \textbf{99} & & \textbf{Total} & \textbf{119} &  \\
\bottomrule
\end{tabular}
\end{adjustbox}
\caption{\textbf{Distribution of Subject  Tasks Across Categories in LAG-MMLU.} In this table we present the frequency of subject-specific tasks categorized into Humanities, STEM, Social Sciences, and Other. This categorization was based on the original MMLU dataset and paper \cite{hendrycks2021measuringmassivemultitasklanguage}}
\label{tab:subject_distribution}
\end{table*}

\newpage

\section{Performance with Standard Errors} 
\label{sec:model_Eval}
% \vspace{-30pt} 

\begin{table*}[h]
\centering
\footnotesize
\setlength{\tabcolsep}{4pt}
\renewcommand{\arraystretch}{1.1}
\resizebox{\textwidth}{!}{%
\begin{tabular}{llcccccc}
\toprule
\textbf{Model} & \textbf{COT}  
  & \multicolumn{2}{c}{\textbf{English (\%)}} 
  & \multicolumn{2}{c}{\textbf{Latvian (\%)}} 
  & \multicolumn{2}{c}{\textbf{Giriama (\%)}} \\
\cmidrule(lr){3-4}\cmidrule(lr){5-6}\cmidrule(lr){7-8}
&  
  & \textbf{0-shot} & \textbf{5-shot}  
  & \textbf{0-shot} & \textbf{5-shot}  
  & \textbf{0-shot} & \textbf{5-shot}  \\
\midrule
\textbf{OpenAI o1} & \xmark  
    & \textbf{92.8} ± 1.2 & -- 
    & \textbf{88.8} ± 1.4 & -- 
    & \textbf{70.8} ± 2.0 & -- \\
\midrule
\multirow{2}{*}{GPT-4o} 
  & \xmark  & 83.6 ± 1.7 & --  
       & 75.6 ± 1.9 & 78.0 ± 1.9  
       & 53.2 ± 2.2 & --  \\
  & \cmark & -- & 84.5 ± 1.6  
       & 79.5 ± 1.8 & --  
       & -- & 62.3 ± 2.2  \\
\midrule
\multirow{2}{*}{Claude 3.5}  
  & \xmark  & 88.8 ± 1.4 & --  
       & 78.2 ± 1.8 & --  
       & 54.8 ± 2.2 & --  \\
  & \cmark & -- & 87.1 ± 1.5  
       & -- & 83.0 ± 1.7  
       & -- & 60.9 ± 2.2  \\
\midrule
\multirow{2}{*}{Gemini 1.5 Pro}  
  & \xmark  & 84.0 ± 1.6 & --  
       & 70.3 ± 2.0 & --  
       & 51.4 ± 2.2 & --  \\
  & \cmark & -- & 85.4 ± 1.6  
       & -- & 79.9 ± 1.8  
       & -- & 61.8 ± 2.2  \\
\midrule
\multirow{2}{*}{DeepSeek-V3}  
  & \xmark  & 87.6 ± 1.5 & --  
       & 70.7 ± 2.0 & --  
       & 49.8 ± 2.2 & --  \\
  & \cmark & -- & 87.9 ± 1.5  
       & -- & 76.1 ± 1.9  
       & -- & 52.3 ± 2.2  \\
\midrule
Llama-70B IT  & \xmark  
    & 81.0 ± 1.8 & --  
    & 57.3 ± 2.2 & --  
    & 41.0 ± 2.2 & --  \\
\midrule
Mistral-large  & \xmark  
    & 80.0 ± 1.8 & --  
    & 49.3 ± 2.2 & --  
    & 35.6 ± 2.1 & --  \\
DeepSeek-R1  & \xmark  
    & 81.4 ± 1.7 & --  
    & 70.2 ± 2.0 & --  
    & 52.2 ± 2.2 & --  \\
\bottomrule
\end{tabular}
}% end resizebox
\caption{\textbf{Performance of Models on LAG-MMLU.} Accuracy (\%) with standard error (SE) across three languages (English, Latvian, and Giriama) under 0-shot and 5-shot settings. Results are presented for models with and without CoT prompting. The LAG-MMLU benchmark reveals a clear performance gap between high-resource and low-resource languages, with OpenAI o1 leading in zero-shot settings across English (92.8\%), Latvian (88.8\%), and Giriama (70.8\%). COT significantly boosts few-shot performance, particularly for GPT-4o, Claude 3.5, and Gemini 1.5 Pro, improving accuracy in Latvian and Giriama. However, Giriama remains challenging for all models, with closed-source models outperforming open ones—though DeepSeek-V3 leads among open models. }
\label{tab:main_results_table2}
\end{table*}

\end{document}